\documentclass[runningheads]{llncs}
\usepackage[T1]{fontenc}
\usepackage[table]{xcolor}  % Include the xcolor package with table option
\usepackage{tabularx}       % Load the tabularx package
\usepackage{graphicx}%
\usepackage{multirow}%
\usepackage{amsmath,amssymb,amsfonts}%
\usepackage{xcolor}%
\usepackage{mathrsfs}%
\usepackage{bm}
\usepackage{capt-of}
\usepackage{pifont}
\newcommand{\cmark}{\text{\ding{51}}}

\usepackage{hyperref}
\usepackage{color}

\usepackage{balance}
\newcommand{\myqed}{\hfill $\blacksquare$}

\definecolor{c0}{RGB}{128, 128, 128}
\definecolor{c1}{RGB}{0, 255, 0}
\definecolor{c2}{RGB}{0, 0, 255}
\definecolor{c3}{RGB}{255, 255, 0}
\definecolor{c4}{RGB}{0, 255, 255}
\definecolor{c5}{RGB}{255, 0, 255}
\definecolor{c6}{RGB}{192, 192, 192}
\definecolor{c7}{RGB}{255, 0, 0}
\definecolor{c8}{RGB}{128, 0, 0}
\definecolor{c9}{RGB}{128, 128, 0}
\definecolor{c10}{RGB}{0, 128, 0}
\definecolor{c11}{RGB}{128, 0, 128}
\definecolor{c12}{RGB}{0, 128, 128}

\definecolor{a_c0}{RGB}{0, 0, 128}
\definecolor{a_c1}{RGB}{0, 128, 128}
\definecolor{a_c2}{RGB}{128, 0, 128}
\definecolor{a_c3}{RGB}{255, 192, 203}
\definecolor{a_c4}{RGB}{0, 255, 0}
\definecolor{a_c5}{RGB}{165, 42, 42}
\definecolor{a_c6}{RGB}{255, 165, 0}
\definecolor{a_c7}{RGB}{245, 245, 220}
\definecolor{a_c8}{RGB}{255, 215, 0}
\definecolor{a_c9}{RGB}{255, 127, 80}

\begin{document}
\title{A Bayesian Approach to Weakly-supervised Laparoscopic Image Segmentation}
\titlerunning{A Bayesian Approach to Weakly-supervised Segmentation}
\author{Zhou Zheng\inst{1}
\and Yuichiro Hayashi\inst{1} 
\and Masahiro Oda\inst{2,1}
\and \\Takayuki Kitasaka\inst{3}
\and Kensaku Mori\inst{1,2,5}}
% index{Zheng, Zhou}
% index{Hayashi, Yuichiro}
% index{Oda, Masahiro}
% index{Kitasaka, Takayuki}
% index{Mori, Kensaku}

\authorrunning{Z. Zheng et al.}
\institute{Graduate School of Informatics, Nagoya University, Nagoya, 464-8601, Aichi, Japan \email{zzheng@mori.m.is.nagoya-u.ac.jp} \\\email{kensaku@is.nagoya-u.ac.jp} \and
Information Technology Center, Nagoya University, Nagoya, 464-8601, Aichi, Japan\and
School of Information Science, Aichi Institute of Technology, \\Toyota, 470-0392, Aichi, Japan\and
Research Center of Medical Bigdata, National Institute of Informatics, \\Chiyoda-ku, 101-8430, Tokyo, Japan}
\maketitle              % typeset the header of the contribution
\begin{abstract}
In this paper, we study weakly-supervised laparoscopic image segmentation with sparse annotations. We introduce a novel Bayesian deep learning approach designed to enhance both the accuracy and interpretability of the model's segmentation, founded upon a comprehensive Bayesian framework, ensuring a robust and theoretically validated method. Our approach diverges from conventional methods that directly train using observed images and their corresponding weak annotations. Instead, we estimate the joint distribution of both images and labels given the acquired data. This facilitates the sampling of images and their high-quality pseudo-labels, enabling the training of a generalizable segmentation model. Each component of our model is expressed through probabilistic formulations, providing a coherent and interpretable structure. This probabilistic nature benefits accurate and practical learning from sparse annotations and equips our model with the ability to quantify uncertainty. Extensive evaluations with two public laparoscopic datasets demonstrated the efficacy of our method, which consistently outperformed existing methods. Furthermore, our method was adapted for scribble-supervised cardiac multi-structure segmentation, presenting competitive performance compared to previous methods. The code is available at \url{https://github.com/MoriLabNU/Bayesian_WSS}.

\end{abstract}
\section{Introduction}
Deep learning-based methods have emerged as a promising solution in laparoscopic image segmentation, which involves model training by using images and the corresponding ground truth~\cite{hong2020cholecseg8k,autolapro}. However, acquiring pixel-wise annotations remains a bottleneck due to the expertise required and the time-consuming annotation process. Thus, it is highly required to explore label-efficient learning for this task. Weakly-supervised segmentation has become an effective paradigm, which takes advantage of sparse annotations such as scribbles, diminishing the reliance on densely annotated labels. A line of promising methods has been proposed in the medical and computer vision communities. For instance, Fuentes-Hurtado et al.~\cite{fuentes2019easylabels} adopted the partial cross-entropy loss (pCE)~\cite{pce} to learn from the labeled pixels while ignoring the unlabeled regions for laparoscopic image segmentation. Luo et al.~\cite{wsl_luo} proposed a dual-branch network and adopted a dynamically mixed pseudo-labels supervision scheme (abbreviated to DBN-DMPLS) for scribble supervision. Liu et al.~\cite{liu2022weakly} introduced an uncertainty-aware self-ensembling and transformation-consistency model (abbreviated to USTM) to learn from limited supervision. Yang et al.~\cite{weakly_spie} presented a method comprising of a graph-model-based scheme, i.e., graph cuts~\cite{graphcuts} and a noisy learning paradigm (abbreviated to GMBM-DLM) for weakly-supervised instrument segmentation. Some studies~\cite{densecrf,gridcrf} investigated penalization terms to regularize training.

Despite the commendable progress made by these methods, they face an array of challenges: (i) the loss of valuable image information, (ii) the error propagation due to generated low-quality supervision signals, and (iii) the limited interpretability and uncertainty quantification. For instance, the study of~\cite{fuentes2019easylabels} is based on the pCE loss~\cite{pce}, only focusing on the labeled pixels while ignoring the unlabeled regions. This selective attention might result in the model overlooking valuable information during training. Some schemes like DBN-DMPLS~\cite{wsl_luo} and USTM~\cite{liu2022weakly} adopt pseudo-label strategies or consistency learning paradigms to generate additional supervision signals. However, the reliability of the generated supervision signals is heavily based on model performance, potentially propagating errors and leading to degraded accuracy. Similarly, GMBM-DLM~\cite{weakly_spie} employs graph cuts to generate pseudo-labels for noisy learning. However, these pseudo-labels, originating from graph cuts, often need to be improved, still carrying a risk of accumulating training errors. Methods such as the combination of the pCE loss with the DenseCRF loss~\cite{densecrf}, and the pairing of the pCE loss with the GridCRF loss~\cite{gridcrf}, integrate a Conditional Random Field (CRF)~\cite{CRF} term within the loss to model the conditional distribution $p\left(\mathbf{y}|\mathbf{x}\right)$, where $\mathbf{x}$ and $\mathbf{y}$ are images and labels, assuming that $\mathbf{y}$ follows a Gibbs distribution. However, these methods do not fully exploit the advantages of a joint probabilistic modeling $p\left(\mathbf{x},\mathbf{y}\right)$, potentially missing out on the richer representation and uncertainty quantification that a fully Bayesian approach offers. Additionally, most existing methods, including those above, do not offer uncertainty estimation for predictions. Given that such models are trained under sparse supervision, assessing the uncertainty of model outputs is essential.

Driven by these perspectives above, we propose a practical and fully Bayesian learning approach for weakly-supervised laparoscopic image segmentation. It is worth mentioning that while the work~\cite{Wang_2022_CVPR}, which inspired our study, presented a fully Bayesian learning method for semi-supervised medical segmentation, utilizing scarce annotations, our research aims to investigate weakly-supervised segmentation. We leverage sparse annotations, leading to a distinct Bayesian formulation tailored to this specific type of supervision. Unlike existing weakly-supervised methods, our method models the joint probability distribution $p\left(\mathbf{x},\mathbf{y}\right)$. By harnessing this joint distribution, our method can generate superior-quality pseudo-labels by accommodating the uncertainties present in these pseudo-labels, thereby reducing the error propagation during training. Besides, our method inherently provides uncertainty estimation for its predictions. Our method provides a more principled approach to handling sparse annotations and enhances the interpretability and reliability of the segmentation results.

Our contributions are mainly threefold: (1) we pioneer rethinking weakly-supervised laparoscopic image segmentation in a Bayesian perspective and propose a novel Bayesian deep learning method for this task, which has a theoretical probabilistic foundation and enhances the accuracy and interpretability of the segmentation results; (2) we extensively validate our method on two public datasets, CholecSeg8k~\cite{hong2020cholecseg8k} and AutoLaparo~\cite{autolapro} and demonstrate its potential solution for this task; and (3) we extend our method to scribble-supervised cardiac multi-structure segmentation~\cite{ACDC,ACDC_scribble} and show its potential for versatility and applicability across different imaging modalities.

\section{Methodology}\label{sec2}
\subsection{Probabilistic modeling}
\textbf{Learning stage.} Generally, given a dataset $\mathcal{D} = \left\{\mathbf{x}, \mathbf{y}\right\}$ containing images $\mathbf{x}$ and the corresponding ground truth $\mathbf{y}$, we can train a model $\mathbf{w}$ in a fully-supervised manner. This procedure also calls modeling posterior distribution $p\left(\mathbf{w}|\mathbf{x},\mathbf{y}\right)$ under the Bayesian framework. However, the dense annotations $\mathbf{y}$ can not be reached in weakly-supervised segmentation. Instead, the weak annotations $\mathbf{y}^{s}$ are provided. The goal is to learn the model $\mathbf{w}$ from the degraded dataset $\mathcal{D}^{s} = \left\{\mathbf{x}, \mathbf{y}^{s}\right\}$, i.e., modeling $p\left(\mathbf{w}|\mathbf{x},\mathbf{y}^{s}\right)$, represented as
\begin{equation}\label{learning_procedure}
p\left(\mathbf{w}|\mathbf{x},\mathbf{y}^{s}\right) = \iiint p\left(\mathbf{w}|\mathbf{x},\mathbf{y}\right) p\left(\mathbf{x},\mathbf{y}|\mathbf{z}\right) p\left(\mathbf{z}|\mathbf{x},\mathbf{y}^{s}\right) \, d\mathbf{x} \, d\mathbf{y} \, d\mathbf{z},
\end{equation}
\noindent where $\mathbf{z}$ are latent variables that determine the joint distribution $p\left(\mathbf{x},\mathbf{y}\right)$, and follow the posterior distribution $p\left(\mathbf{z}|\mathbf{x},\mathbf{y}^{s}\right)$. Given the intractable nature of Eq.~\ref{learning_procedure}, we turn to a Monte Carlo (MC) strategy to approximate $p\left(\mathbf{w}|\mathbf{x},\mathbf{y}^{s}\right)$:
\begin{equation}\label{MC_learning_procedure}
p\left(\mathbf{w}|\mathbf{x},\mathbf{y}^{s}\right) = \frac{1}{MN}\sum_{i=1}^{M} p\left(\mathbf{w}|\mathbf{x}_{\left(i\right)},\mathbf{y}_{\left(i\right)}\right) \sum_{j=1}^{N} p\left(\mathbf{x},\mathbf{y}|\mathbf{z}_{\left(j\right)}\right).  
\end{equation}
\noindent Specifically, we can first sample $\mathbf{z}$ from $p\left(\mathbf{z}|\mathbf{x},\mathbf{y}^{s}\right)$ with $N$ times and then draw $M$ images and the corresponding labels from $p\left(\mathbf{x},\mathbf{y}|\mathbf{z}\right)$ to train the model to obtain approximated $p\left(\mathbf{w}|\mathbf{x},\mathbf{y}^{s}\right)$. Thus, the goal of the learning stage is to obtain $p\left(\mathbf{z}|\mathbf{x},\mathbf{y}^{s}\right)$ and $p\left(\mathbf{x},\mathbf{y}|\mathbf{z}\right)$.

\noindent
\textbf{Inference stage.} After learning $p\left(\mathbf{w}|\mathbf{x},\mathbf{y}^{s}\right)$, given test images $\mathbf{\Bar{x}}$, we can get the corresponding probability maps $\mathbf{\Bar{y}}$ with the following formula:
\begin{equation}\label{inference_procedure}
p\left(\mathbf{\Bar{y}}|\mathbf{\Bar{x}}, \mathbf{x},\mathbf{y}^{s}\right) = \int p\left(\mathbf{\Bar{y}}|\mathbf{\Bar{x}},\mathbf{w}\right) p\left(\mathbf{w}|\mathbf{x},\mathbf{y}^{s}\right)d\mathbf{w}.
\end{equation}
We approximately calculate $p\left(\mathbf{\Bar{y}}|\mathbf{\Bar{x}}, \mathbf{x},\mathbf{y}^{s}\right)$ with MC simulation:
\begin{equation}\label{MC_inference_procedure}
p\left(\mathbf{\Bar{y}}|\mathbf{\Bar{x}}, \mathbf{x},\mathbf{y}^{s}\right) = \frac{1}{T} \sum_{i=1}^{T} p\left(\mathbf{\Bar{y}}|\mathbf{\Bar{x}},\mathbf{w}_{\left(i\right)}\right),
\end{equation}
\noindent where $T$ models are sampled from $p\left(\mathbf{w}|\mathbf{x},\mathbf{y}^{s}\right)$ via MC dropout (MCDP)~\cite{gal2016dropout} with $T$ inference times. By averaging all the probability maps, we can get $\mathbf{\Bar{y}}$ and the related epistemic uncertainty maps $\mathbf{u}$ by calculating the entropy with $-\sum_{c=1}^{C} p\left(\mathbf{\Bar{y}}_{c}|\mathbf{\Bar{x}}, \mathbf{x},\mathbf{y}^{s}\right) \log\left(p\left(\mathbf{\Bar{y}}_{c}|\mathbf{\Bar{x}}, \mathbf{x},\mathbf{y}^{s}\right)\right)$, where $C$ is the number of classes, and $\mathbf{\Bar{y}}_{c}$ denotes the $c$-th channel of $\mathbf{\Bar{y}}$.

\begin{figure}[t]
   \begin{center}
   \includegraphics[width=\textwidth]{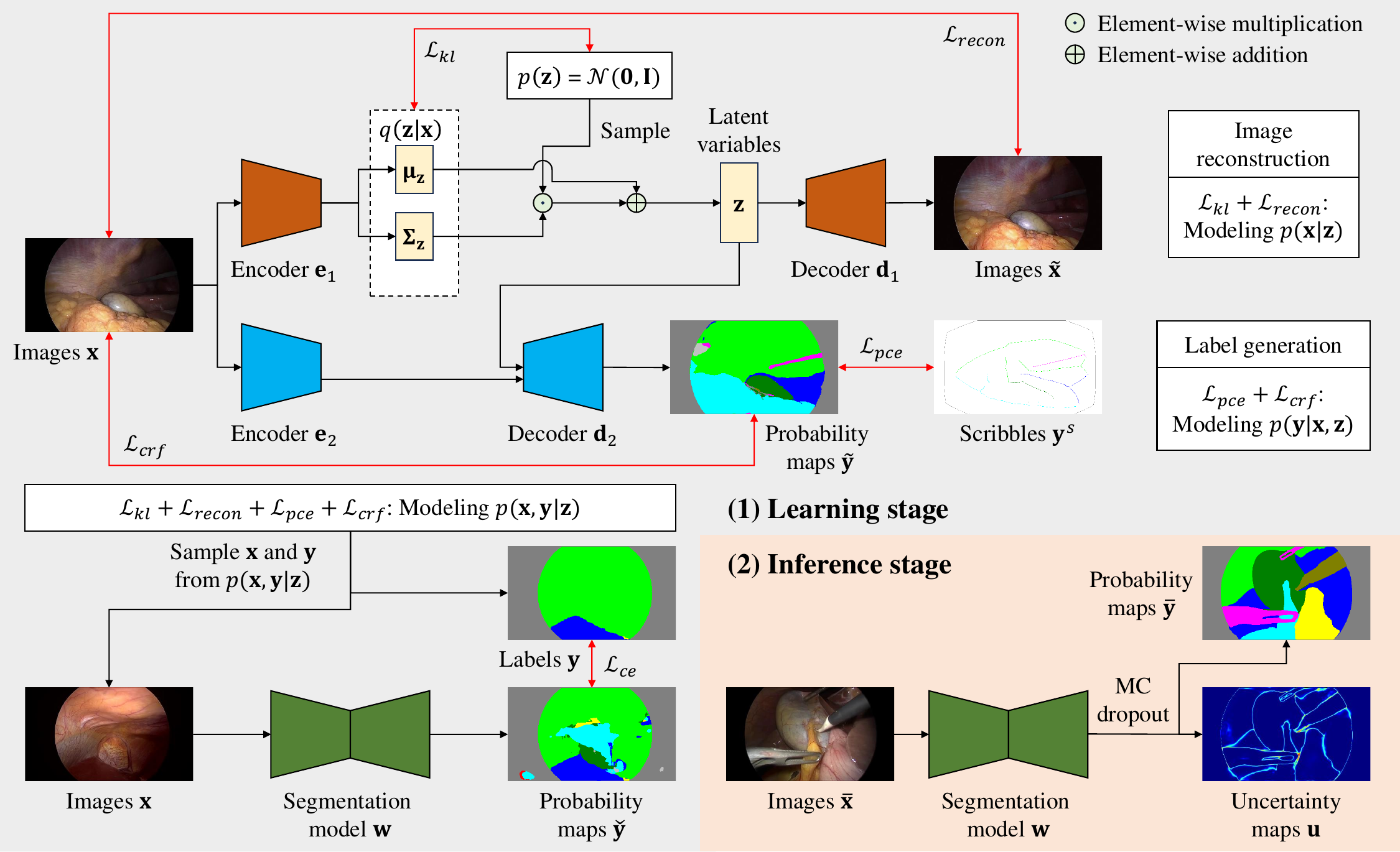}
   \end{center}
   \caption{Flowchart of the proposed framework. At the learning stage, we first learn $p(\mathbf{x},\mathbf{y}|\mathbf{z})$ by modeling $p(\mathbf{x}|\mathbf{z})$ for image reconstruction and $p(\mathbf{y}|\mathbf{x},\mathbf{z})$ for label generation. After obtaining $p(\mathbf{x},\mathbf{y}|\mathbf{z})$, we 
   sample pairs of $\mathbf{x}$ and $\mathbf{y}$ from $p(\mathbf{x},\mathbf{y}|\mathbf{z})$ to learn a segmentation model, i.e., $p(\mathbf{w}|\mathbf{x},\mathbf{y})$. At the inference stage, we obtain the prediction and corresponding epistemic uncertainty estimation with MC dropout.}\label{flowchart}
\end{figure}

\subsection{Overview of proposed framework}
The flowchart of our framework is shown in Fig.~\ref{flowchart}. The learning and inference procedures are guided by the formulation presented in Eq.~\ref{learning_procedure} and~\ref{inference_procedure}. 

We begin with the premise that $\mathbf{z}$ are statistically independent from $\mathbf{x}$ and $\mathbf{y}^{s}$, allowing us to simplify $p\left(\mathbf{z}|\mathbf{x},\mathbf{y}^{s}\right)$ to the prior distribution $p\left(\mathbf{z}\right)$. This assumption sets the stage for the initial step of our method, which involves the derivation of the Evidence Lower Bound (ELBO) as shown in Eq.~\ref{derivation}, with its proof provided in the supplementary material. Eq.~\ref{derivation} facilitates the decomposition of our target log-likelihood $\log p\left(\mathbf{x}, \mathbf{y}\right)$ into manageable components, allowing us to introduce a variational distribution $q\left(\mathbf{z}|\mathbf{x}\right)$ that approximates the intractable prior distribution $p\left(\mathbf{z}\right)$. Our objective is to maximize the ELBO, expressed as in Eq.~\ref{ELBO}.
\begin{equation}\label{derivation}
\begin{aligned}
\log p\left(\mathbf{x}, \mathbf{y}\right) &\geq \mathbb{E}_{\mathbf{z} \sim q}\left[\log p\left(\mathbf{y}|\mathbf{x}, \mathbf{z}\right) + \log p\left(\mathbf{x}|\mathbf{z}\right)\right] - \mathbb{E}_{\mathbf{z} \sim q}\left[\log \frac{q\left(\mathbf{z}|\mathbf{x}\right)}{p\left(\mathbf{z}\right)}\right].
\end{aligned}
\end{equation}
\begin{equation}\label{ELBO}
\text{ELBO} = \mathbb{E}_{\mathbf{z} \sim q}\left[\log p\left(\mathbf{y}|\mathbf{x}, \mathbf{z}\right)\right] + \mathbb{E}_{\mathbf{z} \sim q}\left[\log p\left(\mathbf{x}|\mathbf{z}\right)\right] - \text{KL}\left[q\left(\mathbf{z}|\mathbf{x}\right) || p\left(\mathbf{z}\right)\right].
\end{equation}
\noindent
\textbf{Latent space modeling.}
We follow the conditional variational auto-encoder (CVAE)~\cite{CVAE} to modulate the latent variables $\mathbf{z}$ with the input images $\mathbf{x}$, formalizing the distribution $q\left(\mathbf{z}|\mathbf{x}\right)$. Specifically, an encoder $\mathbf{e}_{1}$ is applied to map $\mathbf{x}$ to the latent space. Following the common settings, we assume $q\left(\mathbf{z}|\mathbf{x}
\right) = \mathcal{N}\left(\bm{\mu}_{\mathbf{z}},\mathbf{\Sigma}_{\mathbf{z}}\right)$ is a multivariate Gaussian distribution, and $p\left(\mathbf{z}\right)$ is a multivariate standard normal distribution $\mathcal{N}\left(\textbf{0},\textbf{I}\right)$. We calculate a Kullback–Leibler (KL) divergence loss $\mathcal{L}_{kl}\left(q\left(\mathbf{z}|\mathbf{x}\right), p\left(\mathbf{z}\right)\right)$ to push $q\left(\mathbf{z}|\mathbf{x}\right)$ closer to $\mathcal{N}(\textbf{0},\textbf{I})$. On the other hand, a decoder $\mathbf{d}_{1}$ is employed to reconstruct the images $\mathbf{\Tilde{x}}$ from the sampled latent representation\footnote{In practice, $\mathbf{z}$ undergoes several necessary transformations between $\mathbf{e}_{1}$ and $\mathbf{d}_{1}$. Details of the network configuration for this part are given in the supplementary material.}, creating a cycle that optimizes the reconstruction likelihood $p\left(\mathbf{x}|\mathbf{z}\right)$ with a loss $\mathcal{L}_{recon}\left(\mathbf{\Tilde{x}},\mathbf{x}\right)$ that calculates the mean square error (MSE).

\noindent
\textbf{Conditional random field modeling.}
We adopt the CRF~\cite{CRF}, which is characterized by a Gibbs distribution, to maximize $p\left(\mathbf{y}|\mathbf{x}, \mathbf{z}\right)$. Numerous studies like~\cite{densecrf,gridcrf} have explored the integration of CRF into training. We use the pairing of the pCE loss~\cite{pce} with the DenseCRF loss~\cite{densecrf} to model CRF.

In practice, we introduce another encoder $\mathbf{e}_{2}$ to encode the images $\mathbf{x}$ to high-level features and concatenate them with the latent variables $\mathbf{z}$ together and use another decoder $\mathbf{d}_{2}$ to obtain the predicted probability maps $\mathbf{\Tilde{y}}$. We make use of the sparse annotations $\mathbf{y}^{s}$ to optimize $\mathbf{e}_{2}$ and $\mathbf{d}_{2}$ to generate sub-optimal predictions by calculating the pCE loss $\mathcal{L}_{pce}\left(\mathbf{\Tilde{y}},\mathbf{y}^{s}\right)$ between $\mathbf{\Tilde{y}}$ and $\mathbf{y}^{s}$:
\begin{equation}
\mathcal{L}_{pce}\left(\mathbf{\Tilde{y}}, \mathbf{y}^{s}\right)  =  - \sum_{a\in \Omega^{s}}\sum_{c=1}^{C} \mathbf{y}^{s}_{a,c}\log\left(\mathbf{\Tilde{y}}_{a,c}\right),
\end{equation}
\noindent where $\Omega^{s}$ represents the set of indices corresponding to the pixels with sparse annotations. Meanwhile, we incorporate the DenseCRF loss $\mathcal{L}_{crf}\left(\mathbf{\hat{y}}\right)$:
\begin{equation}
\mathcal{L}_{crf}\left(\mathbf{\hat{y}}\right) =  \sum_{c=1}^{C} \mathbf{\hat{y}}_{c}^{\prime}\mathbf{K} \left(\textbf{1} - \mathbf{\hat{y}}_{c}\right),
\end{equation}
\noindent where $\mathbf{\hat{y}}_{c}$ is a vector associated with class $c$, containing all elements $\mathbf{\Tilde{y}}_{a,c}$ from $\mathbf{\Tilde{y}}_{c}$ for $a \in \Omega$, where $\Omega$ denotes the set of indices of all pixels in $\mathbf{\Tilde{y}}_{c}$, and $\mathbf{\Tilde{y}}_{a,c}$ is the component of the $a$-th pixel in the $c$-th channel of $\mathbf{\Tilde{y}}$. Besides, $\mathbf{K}$ is a matrix of pairwise discontinuity costs. Each element $k_{a,b}$ in $\mathbf{K}$ is determined by a dense Gaussian kernel~\cite{densecrf}. By optimizing both $\mathcal{L}_{pce}\left(\mathbf{\Tilde{y}}, \mathbf{y}^{s}\right)$ and $\mathcal{L}_{crf}\left(\mathbf{\hat{y}}\right)$ to maximize $p(\mathbf{y}|\mathbf{x}, \mathbf{z})$, we can generate high-quality pseudo-labels and treat them as unobserved labels $\mathbf{y}$.

\noindent
\textbf{Training procedures.}
Firstly, maximizing ELBO in Eq.~\ref{ELBO} is equivalent to optimizing the following training objective:
\begin{equation}\label{training_loss}
\mathcal{L}_{ELBO} = \mathcal{L}_{pce}\left(\mathbf{\Tilde{y}}, \mathbf{y}^{s}\right) + \alpha \mathcal{L}_{kl}\left(q\left(\mathbf{z}|\mathbf{x}\right), p\left(\mathbf{z}\right)\right) + \beta
\mathcal{L}_{recon}\left(\mathbf{\Tilde{x}}, \mathbf{x}\right) + \gamma \mathcal{L}_{crf}\left(\mathbf{\hat{y}}\right),
\end{equation}

\noindent where $\alpha$, $\beta$, and $\gamma$ are weighting coefficients. It is important to note that while our work shares the same ELBO as the work~\cite{Wang_2022_CVPR}, our method's optimization target and loss function are distinct. The main difference is that our method assumes $p\left(\mathbf{y}|\mathbf{x}, \mathbf{z}\right)$ follows a Gibbs distribution and maximizes it with the CRF.

By optimizing $\mathcal{L}_{ELBO}$, we can obtain the joint distribution $p\left(\mathbf{x},\mathbf{y}\right)$ and draw pairs of $\mathbf{x}$ and $\mathbf{y}$. In the practical implementation, to make use of the scribble annotations $\mathbf{y}^{s}$, we merge $\mathbf{y}^{s}$ into the generated labels $\mathbf{y}$. Concretely, we create binary masks $\mathbf{\Gamma}$ based on $\mathbf{y}^{s}$, where the value of 0 indicates the labeled region, and 1 represents the unlabeled region in $\mathbf{y}^{s}$. The final labels, y, are then generated by $\mathbf{y} = \left(1 - \mathbf{\Gamma}\right) \odot \mathbf{y}^{s} + \mathbf{\Gamma} \odot \mathbf{y}$, where $\odot$ denotes element-wise multiplication. We utilize $\mathbf{x}$ and $\mathbf{y}$ to train a model $\mathbf{w}$ with a cross-entropy (CE) loss $\mathcal{L}_{ce}\left(\mathbf{\check{y}},\mathbf{y}\right)$ between the predicted probability maps $\mathbf{\check{y}}$ and labels $\mathbf{y}$.

\noindent
\textbf{Inference stage.} Upon completing the training of $\mathbf{w}$, for the test images $\mathbf{\Bar{x}}$, we utilize MCDP~\cite{gal2016dropout} to obtain the probability maps $\mathbf{\Bar{y}}$ and the corresponding epistemic uncertainty maps $\mathbf{u}$ with $T$ inference times.

\section{Experiments and results}\label{sec3}
\subsection{Experimental setup}
\textbf{Datasets.} Our method was validated using two public laparoscopic datasets: CholecSeg8k~\cite{hong2020cholecseg8k} and AutoLaparo~\cite{autolapro}. The CholecSeg8k dataset~\cite{hong2020cholecseg8k} collects 8080 images from the public dataset Cholec80~\cite{twinanda2016endonet} and provides the corresponding ground truth of 13 classes. The resolution of each image is $854 \times 480$ pixels. The AutoLaparo dataset~\cite{autolapro} offers three tasks. We focused on Task 3 of segmentation. This task includes 1800 frames annotated across 10 classes. Each frame is in $1920 \times 1080$ pixels. The dataset was split into training, validation, and test sets with 1020, 342, and 438 frames, respectively, following the official division.

\noindent
\textbf{Weak label generation.}
The CholecSeg8k and AutoLaparo datasets do not provide weak annotations. Inspired by previous works~\cite{fuentes2019easylabels,ACDC_scribble,weakly_spie,GAO2022102515} that extracted skeletons from the ground truth to generate weak annotations, we obtained sparse annotations with the skeletonization method~\cite{zhang1984fast} in our study.

\noindent
\textbf{Evaluation metrics.} We employed the Dice score (DC) [\%], Jaccard (JA) [\%], sensitivity (SE) [\%], and specificity (SP) [\%] as metrics.

\noindent
\textbf{Implementation details.}
We leveraged the U-Net~\cite{unet} augmented with dropout layers as the backbone. Specifically, $\mathbf{w}$ adopted the U-Net backbone. The elements $\mathbf{e}_{1}$ and $\mathbf{e}_{2}$ functioned as variants of U-Net's encoder, while $\mathbf{d}_{1}$ and $\mathbf{d}_{2}$ were derivatives of U-Net's decoder. The dimension of $\mathbf{z}$ was set to 256. For hyper-parameters in Eq.~\ref{training_loss}, we empirically determined a combination of $\alpha = 10^{-3}$, $\beta = 10^{-1}$, and $\gamma = 10^{-8}$. We set $N$ to 5 and $T$ to 15 (see section~\ref{sec_ablation}). More details on implementation are given in the supplementary material.

\noindent
\textbf{Baselines.} The upper-bound result was obtained with fully supervised segmentation (denoted as Fully-sup), while the lower-bound performance was yielded by training the model with sparse annotations via the pCE loss~\cite{pce}. We further implemented four state-of-the-art (SOTA) methods for comparison: DBN-DMPLS~\cite{wsl_luo}, USTM~\cite{liu2022weakly}, GMBM-DLM~\cite{weakly_spie}, and a combination of the pCE loss with DenseCRF losses (notated as pCE+DenseCRF)~\cite{densecrf}. Evaluations were conducted using 5-fold cross-validation on the CholecSeg8k dataset and repeated in 5 trials on the AutoLaparo dataset.

\begin{table}[t]
  \centering
    \caption{Quantitative comparison of various methods
on the CholecSeg8k dataset and AutoLaparo test set. Experiments were conducted with 5-fold cross-validation on the
CholecSeg8k dataset and repeated in 5 trials on the AutoLaparo dataset.}\label{quantitative_results}
    \resizebox{0.6\textwidth}{!}{
        \begin{tabular}{l|cccc|cccc}
\hline
\multirow{2}{*}{Method} & \multicolumn{4}{c|}{CholecSeg8k} & \multicolumn{4}{c}{AutoLaparo} \\ \cline{2-9}
      & DC & JA & SE & SP & DC & JA & SE & SP \\ \hline
pCE~\cite{pce}        &    51.6         & 46.5    &  53.4           &  98.9   &    25.9         & 20.9    &  24.9           &  94.4  \\ \hline
DBN-DMPLS~\cite{wsl_luo} &    61.3         & 56.1    &     62.2        &   98.8   &    28.3         & 22.9    &  27.0           &  94.6     \\ \hline
USTM~\cite{liu2022weakly} &     64.9        &  58.7   &     66.3        &   99.1    &    26.5         & 21.3    &  25.2           &  94.4   \\ \hline
GMBM-DLM~\cite{weakly_spie} &    78.5         & 71.7    &   76.9          &  99.0    &    28.0         & 23.2    &  24.8           &  94.1    \\ \hline
pCE+DenseCRF~\cite{densecrf} &     76.6        & 70.4    &       76.8      &    \textbf{99.4}   &    28.2         & 22.9    &  25.8           &  94.5    \\ \hline
pCE+DenseCRF$^{\dag}$~\cite{densecrf} &     78.8        & 72.3    &       78.5      &    \textbf{99.4}   &    29.6         & 24.0    &  26.7           &  94.2    \\ \hline
Ours w/o MCDP &    80.9         & 74.3     &      80.6        &        \textbf{99.4}     &    32.1         & 26.2    &  28.8           &  94.6    \\ \hline
Ours &    \textbf{82.3}         & \textbf{75.9}    &      \textbf{82.2}       &        \textbf{99.4}     &    \textbf{33.4}         & \textbf{27.5}    &  \textbf{29.9}           &  \textbf{94.9}    \\ \hline \hline
Fully-sup        &    88.7         &  83.7   & 88.8            &   99.6    &    37.4         & 32.2    &  36.2           &  95.9    \\ \hline 
\end{tabular}
}
\end{table}

\subsection{Experimental results}
Table~\ref{quantitative_results} presents quantitative comparisons of various methods on the CholecSeg8k dataset and AutoLaparo test set. All methods performed better than the pCE loss, demonstrating their efficacy in learning from weak labels. Notably, our method consistently surpassed other SOTAs by a large margin, reaching closer to the upper-bound accuracy, particularly in DC, JA, and SE.

In addition, one might consider a straightforward approach, referred to as pCE+DenseCRF$^{\dag}$\footnote{pCE+DenseCRF$^{\dag}$ initially trains a model with pCE+DenseCRF by modeling the conditional distribution $p\left(\mathbf{y}|\mathbf{x}\right)$, then uses this model to generate pseudo-labels and treat them as unobserved labels $\mathbf{y}$, and finally retrains a new model with $\mathbf{x}$ and $\mathbf{y}$.}, which bears similarity to our approach but omits the latent variables $\mathbf{z}$. However, pCE+DenseCRF$^{\dag}$ fails to capture the uncertainty in the pseudo-label generation process. In contrast, by conditioning on $\mathbf{z}$, our method leverages sampling of $\mathbf{z}$ to incorporate uncertainty into the model of $p\left(\mathbf{x},\mathbf{y}\right)$, thereby improving the pseudo-label quality in practice. To prove this, we compared our method with pCE+DenseCRF$^{\dag}$. As shown in Table~\ref{quantitative_results}, while pCE+DenseCRF$^{\dag}$ generally improved the accuracy over pCE+DenseCRF, it still fell short of the performance achieved by ours with and without MCDP, underscoring the superiority of our method in generating superior pseudo-labels.

\subsection{Ablation studies}\label{sec_ablation}
Ablation studies were conducted using the CholecSeg8k dataset. These studies were divided into three stages. Initially, we evaluated each loss component's contribution by adding them incrementally with fixed sample time $N = 1$ and inference time $T = 1$ (without MCDP). Next, keeping all loss components and fixing $T=1$ (without MCDP), we identified the optimal $N$. Finally, maintaining all loss components and the optimal $N$, we activated MCDP to assess the effects of different inference times $T$ ($T>1$). Detailed results are presented in Table~\ref{ablation_study}.

\noindent
\textbf{Efficacy of loss components.}
Using only $\mathcal{L}_{pce}$ yielded the lowest performance. Sequentially adding $\mathcal{L}_{kl}$, $\mathcal{L}_{recon}$, and $\mathcal{L}_{crf}$ improved results, confirming each loss's contribution. Notably, including $\mathcal{L}_{recon}$ largely enhanced accuracy, underscoring its importance in generating high-quality pseudo-labels.

\noindent
\textbf{Influence of sample time $N$.} $N =3$ obtained the best results compared to other tested values (1, 5, and 7), leading us to set $N$ to 3 for our experiments.

\noindent
\textbf{Influence of inference time $T$.} By sampling model weights using MCDP at 5, 10, 15, and 20 times, we found that MCDP generally improved the accuracy. As suggested by the results, we set $T$ to 15 for all experiments.

\begin{figure}[t]
\centering
\begin{minipage}[t]{0.665\textwidth}
    \captionof{table}{Ablation studies on efficacy of loss components, influence of sample time $N$, and impact of inference time $T$ with the CholecSeg8k dataset.}\label{ablation_study}
\resizebox{\textwidth}{!}{
\begin{tabular}{cccc|c|c|cccc}
\hline
\multicolumn{4}{c|}{Loss component} & Sample time & Inference time & \multicolumn{4}{c}{Metrics} \\ \hline
$\mathcal{L}_{pce}$  & $\mathcal{L}_{kl}$  & $\mathcal{L}_{recon}$  & $\mathcal{L}_{crf}$ & $N$                & $T$                   & DC     & JA     & SE     & SP    \\ \hline
 \cmark    &        &     &     & 1                & 1                   &   52.7     &   47.0     &   55.1     &  98.8     \\
 \cmark    &  \cmark      &     &     & 1                & 1                   &  57.3      &   51.5     &   58.9     &   99.0    \\
 \cmark    &   \cmark     & \cmark    &     & 1                & 1                   &   57.9     & 52.0       &  60.0      &  98.9       \\
 \cmark    &   \cmark     &  \cmark   &  \cmark   & 1                & 1                   &   \textbf{78.9}     & \textbf{72.6}       &  \textbf{78.6}      &  \textbf{99.4}     \\ \hline
  \cmark    &   \cmark     &  \cmark   &  \cmark   & 1                & 1                   &   78.9     & 72.6       &  78.6      &  \textbf{99.4}  \\
  \cmark   &    \cmark    &  \cmark   &  \cmark   & 3                & 1                   &   \textbf{80.9}     &  \textbf{74.3}      &   \textbf{80.6}     &  \textbf{99.4}     \\
 \cmark    &  \cmark      &  \cmark   &  \cmark   & 5                & 1                   &   79.9     &   73.4     &  80.0      &  \textbf{99.4}     \\
 \cmark    &    \cmark    &  \cmark   &  \cmark   & 7                & 1                   &   79.0     &   72.6     &   78.8     &  \textbf{99.4}     \\ \hline
 \cmark    &  \cmark      &  \cmark   &  \cmark   & 3                & 1                   &   80.9     &   74.3     &  80.6      &  \textbf{99.4}      \\
  \cmark   &   \cmark     &  \cmark   & \cmark    &  3                & 5                   &   82.1     &    75.7    &    82.0    &  \textbf{99.4}     \\
 \cmark    &   \cmark     &  \cmark   &  \cmark   &  3                & 10                  &   82.2     &  75.8      &  82.1      &   \textbf{99.4}    \\
 \cmark    &   \cmark     &  \cmark   &  \cmark   &   3               & 15                  &   \textbf{82.3}     &  \textbf{75.9}      &  \textbf{82.2}      &   \textbf{99.4}    \\
  \cmark   &   \cmark     &  \cmark   &  \cmark   &  3                & 20                  &  82.2      &   \textbf{75.9}     &  \textbf{82.2}      & \textbf{99.4}      \\ \hline
\end{tabular}
}
\end{minipage}
\hfill
\begin{minipage}[t]{0.325\textwidth}
    \captionof{table}{Quantitative comparison of segmentation performance on the ACDC dataset. Other results are adopted from~\cite{wsl_luo}.}\label{ACDC_results}
        \resizebox{\textwidth}{!}{
\begin{tabular}{l|cc}
\hline
Method      & DC       & 95HD       \\ \hline
pCE~\cite{pce}    & 68.6       & 173.3      \\ \hline
RW~\cite{grady2006random}          & 78.8       & 10.0       \\\hline
USTM~\cite{liu2022weakly}        & 78.6       & 102.2      \\\hline
S2L~\cite{scribble2label}          & 83.2       & 38.9       \\\hline
MLoss~\cite{kim2019mumford}         & 83.9       & 27.7       \\\hline
EM~\cite{EM}           & 84.6       & 39.0       \\\hline
RLoss~\cite{pce}          & 85.6       & \textbf{6.9}        \\\hline
DBN-DMPLS~\cite{wsl_luo}         & 87.2       & 9.9        \\ \hline
Ours       & \textbf{87.5}       & 9.0       \\ \hline \hline
Fully-sup  & 89.8       & 7.0        \\\hline
\end{tabular}
}
  \end{minipage}

\end{figure}

\subsection{Extension to other domains}
We further explored the generalizability of our method by applying it to cardiac multi-structure segmentation using the ACDC dataset~\cite{ACDC} and corresponding scribble annotations~\cite{ACDC_scribble}. This dataset includes 200 cine-MRI volumes from 100 patients, along with the ground truth for the right ventricle, myocardium, and left ventricle. We adopted the 2D U-Net model~\cite{unet} as the backbone. The implementation details are presented in the supplementary material.

Quantitative results of DC and 95\% Hausdorff distance (95HD) [mm] are summarized in Table~\ref{ACDC_results}. We referenced results of existing methods reported in~\cite{wsl_luo} for comparison purposes, considering that the same U-Net backbone and 5-fold cross-validation splitting were used. Our method showed competitive results compared to previous models, underlining its potential generalizability to different medical imaging domains.

\section{Discussion and conclusions}\label{sec4}
We proposed a novel method grounded in a fully Bayesian learning paradigm for weakly-supervised laparoscopic image segmentation. Extensive evaluations have demonstrated our method's potential solution to this task and its adaptability to different imaging modalities.

A primary limitation of our method is high computational demand. Future efforts will aim to lower computational expenses. Moreover, we simulated the sparse annotations due to the lack of real weak labels, inspired by previous works~\cite{fuentes2019easylabels,ACDC_scribble,weakly_spie,GAO2022102515}. Thus, applying our method to more datasets with real weak labels, which more closely mirror real-world scenarios, is an aspect of future work. Additionally, we urge both ourselves and the community to contribute datasets featuring real weak labels to facilitate continued studies in this area.

\subsubsection*{Acknowledgments.}
This work was supported by JSPS KAKENHI (24H00720, 24K03262), JST CREST (JPMJCR20D5), JST [Moonshot R\&D] (JPMJMS2033, JPMJMS2214), and JSPS Bilateral Joint Research Project.

\subsubsection*{Disclosure of Interests.}
The authors have no competing interests to declare that are relevant to the content of this article.

\bibliographystyle{splncs04}
\bibliography{Bayesian_WSS_MICCAI2024_arxiv}

\clearpage

\section*{Supplementary Material}

\begin{proof}
As for $p\left(\mathbf{x}, \mathbf{y}\right)$, we have:
\begin{equation}
\begin{aligned}
\log p\left(\mathbf{x}, \mathbf{y}\right) &= \log \int_{\mathbf{z}} p\left(\mathbf{x}, \mathbf{y}, \mathbf{z}\right) \, d\mathbf{z} = \log \int_{\mathbf{z}} \frac{p\left(\mathbf{x}, \mathbf{y}, \mathbf{z}\right) q\left(\mathbf{z}|\mathbf{x}\right)}{q\left(\mathbf{z}|\mathbf{x}\right)} \, d\mathbf{z} \\
&\geq \mathbb{E}_{\mathbf{z} \sim q}\left[\log \frac{p\left(\mathbf{x}, \mathbf{y}, \mathbf{z}\right)}{q\left(\mathbf{z}|\mathbf{x}\right)}\right] = \mathbb{E}_{\mathbf{z} \sim q} \left[\log \frac{p\left(\mathbf{y} | \mathbf{x}, \mathbf{z}\right) p\left(\mathbf{x} | \mathbf{z}\right) p\left(\mathbf{z}\right)}{q\left(\mathbf{z}|\mathbf{x}\right)}\right] \\
&=\mathbb{E}_{\mathbf{z} \sim q}\left[\log p\left(\mathbf{y}|\mathbf{x}, \mathbf{z}\right) + \log p\left(\mathbf{x}|\mathbf{z}\right)\right] - \mathbb{E}_{\mathbf{z} \sim q}\left[\log \frac{q\left(\mathbf{z}|\mathbf{x}\right)}{p\left(\mathbf{z}\right)}\right],
\end{aligned}
\end{equation}\end{proof}

\noindent
where $q\left(\mathbf{z}|\mathbf{x}\right)$ is a variational distribution, and $\mathbb{E}_{\mathbf{z} \sim q}$ denotes the expectation over $q\left(\mathbf{z}|\mathbf{x}\right)$. We finish the proof by deriving the ELBO in Eq.~(5). \myqed

\begin{table}[!htb]
   \caption{Implementation details. Experiments were performed on PyTorch.}
   \centering
   \resizebox{\textwidth}{!}{
\begin{tabular}{p{2.7cm}|p{4cm}|p{4cm}|p{4cm}}
\hline
Dataset         & CholecSeg8k      & AutoLaparo    & ACDC            \\ \hline\hline
Backbone    & U-Net & U-Net & U-Net \\ \hline
Preprocessing    & Resized each image to $432 \times 240$ pixels and normalized the intensities to [0,1]  & Resized each image to $480 \times 240$ pixels and normalized the intensities to [0,1]    & Resized each slice to $256 \times 256$ pixels and normalized the intensities to [0,1] \\ \hline
Input size & $432 \times 240$ & $480 \times 240$ & $256 \times 256$ \\ \hline
Optimizer       & Adam with a weight decay of $10^{-4}$        & Adam with a weight decay of $10^{-4}$    & SGD with a weight decay of $10^{-4}$ and a momentum of 0.9  \\ \hline
Batch size      & 8                           & 8    &  8  \\ \hline
Training epochs or iterations &  1st stage: $\mathbf{e}_{1}$, $\mathbf{d}_{1}$, $\mathbf{e}_{2}$, and $\mathbf{d}_{2}$ were jointly trained for 100 epochs, \newline 2nd stage: $\mathbf{w}$ was trained for 100 epochs           & 1st stage: $\mathbf{e}_{1}$, $\mathbf{d}_{1}$, $\mathbf{e}_{2}$, and $\mathbf{d}_{2}$ were jointly trained for 200 epochs, \newline 2nd stage: $\mathbf{w}$ was trained for 200 epochs & 1st stage: $\mathbf{e}_{1}$, $\mathbf{d}_{1}$, $\mathbf{e}_{2}$, and $\mathbf{d}_{2}$ were jointly trained for 90000 iterations, \newline 2nd stage: $\mathbf{w}$ was trained for 90000 iterations    \\ \hline
Learning rate   &    1st stage: $10^{-4}$, \newline 2nd stage: $10^{-4}$            & 1st stage: $10^{-4}$, \newline 2nd stage: $10^{-4}$     & 1st stage: $10^{-2}\times(1-\eta/90000)^{0.9}$, \newline 2nd stage: $10^{-2}\times(1-\eta/90000)^{0.9}$, \newline $\eta$ is the current iteration  \\ \hline
Dimension of $\mathbf{z}$    & 256 & 256 & 256 \\ \hline
$\alpha$    & $10^{-3}$ & $10^{-3}$ & $10^{-3}$ \\ \hline
$\beta$    & $10^{-1}$ & $10^{-1}$ & $10^{-1}$ \\ \hline
$\gamma$    & $10^{-8}$ & $10^{-8}$ & $10^{-8}$ \\ \hline
$N$    & 3 & 3 & 3 \\ \hline
$T$    & 15 & 15 & 15 \\ \hline
Execution manner    & 5-fold cross validation & 5-trial repeats  & 5-fold cross validation \\ \hline
\end{tabular}
}
\end{table}

\begin{figure}[!htb]
   \begin{center}
   \includegraphics[width=\textwidth]{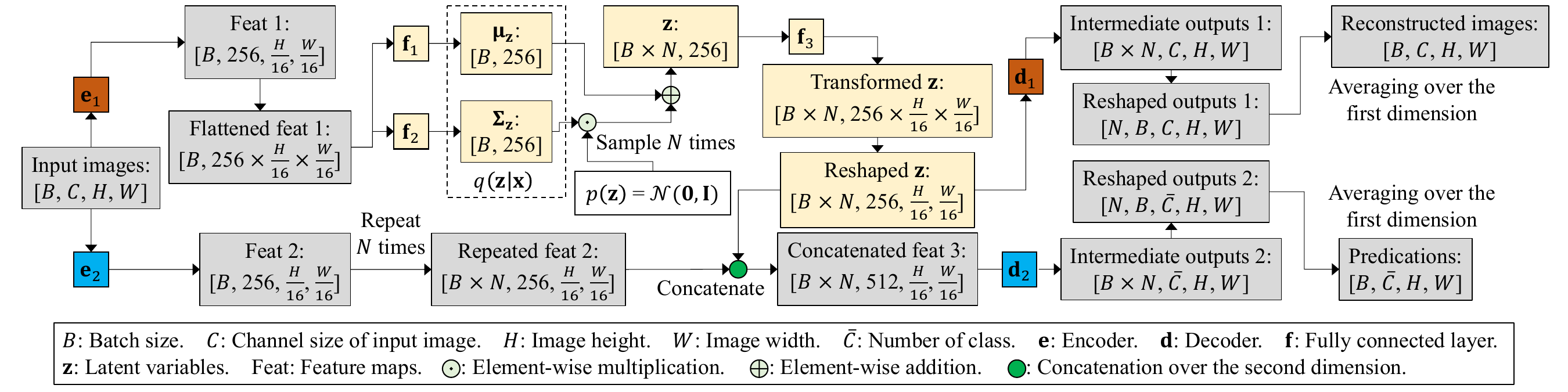}
   \end{center}
   \caption{Network configuration for modeling $p(\mathbf{x},\mathbf{y}|\mathbf{z})$. For simplicity, specifics of the encoder and decoder layers are excluded, and skip connections are omitted.}
\end{figure}

\begin{figure}[!htb]
  \centering
    \begin{minipage}{0.35\textwidth}
    \captionof{table}{Illustration of the CholecSeg8k.}
        \resizebox{\textwidth}{!}{
      \begin{tabular}{|l|l|l|}
        \hline
        Class ID & Object                 & Color \\ \hline
        Class 0       & Black background       &\cellcolor{c0} \\ \hline
        Class 1       & Abdominal wall         &\cellcolor{c1} \\ \hline
        Class 2       & Liver                  &\cellcolor{c2} \\ \hline
        Class 3       & Gastrointestinal tract &\cellcolor{c3} \\ \hline
        Class 4       & Fat                    &\cellcolor{c4} \\ \hline
        Class 5       & Grasper                &\cellcolor{c5} \\ \hline
        Class 6       & Connective tissue      &\cellcolor{c6} \\ \hline
        Class 7       & Blood                  &\cellcolor{c7} \\ \hline
        Class 8       & Cystic duct            &\cellcolor{c8} \\ \hline
        Class 9       & L-hook electrocautery  &\cellcolor{c9} \\ \hline
        Class 10      & Gallbladder            &\cellcolor{c10} \\ \hline
        Class 11      & Hepatic vein           &\cellcolor{c11} \\ \hline
        Class 12      & Liver ligament         &\cellcolor{c12} \\ \hline
      \end{tabular}
      }
  \end{minipage}
  \hfill
  \begin{minipage}{0.35\textwidth}
    \captionof{table}{Illustration of the AutoLaparo. I: Instrument}
    \resizebox{\textwidth}{!}{
      \begin{tabular}{|l|l|l|}
        \hline
        Class ID & Object                 & Color \\ \hline
        Class 0       & Background       &\cellcolor{a_c0} \\ \hline
        Class 1       & Manipulation of I-1         &\cellcolor{a_c1} \\ \hline
        Class 2       & Shaft of I-1                  &\cellcolor{a_c2} \\ \hline
        Class 3       & Manipulation of I-2 &\cellcolor{a_c3} \\ \hline
        Class 4       & Shaft of I-2                    &\cellcolor{a_c4} \\ \hline
        Class 5       & Manipulation of I-3                &\cellcolor{a_c5} \\ \hline
        Class 6       & Shaft of I-3      &\cellcolor{a_c6} \\ \hline
        Class 7       & Manipulation of I-4                  &\cellcolor{a_c7} \\ \hline
        Class 8       & Shaft of I-4            &\cellcolor{a_c8} \\ \hline
        Class 9       & Uterus  &\cellcolor{a_c9} \\ \hline
      \end{tabular}
      }
  \end{minipage}%
  \hfill
    \begin{minipage}{0.29\textwidth}
  \resizebox{\textwidth}{!}{\includegraphics[width=\textwidth]{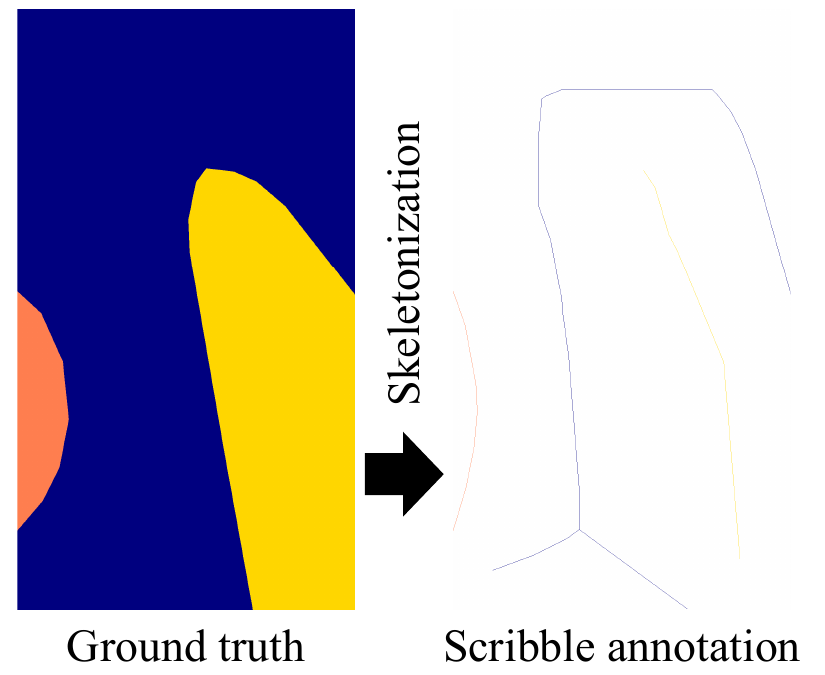}}
    \caption{An example of weak annotation simulation with skeletonization. The white area indicates unlabeled region.}
  \end{minipage}%
\end{figure}

\begin{figure}[!htb]
   \begin{center}
   \includegraphics[width=0.65\textwidth]{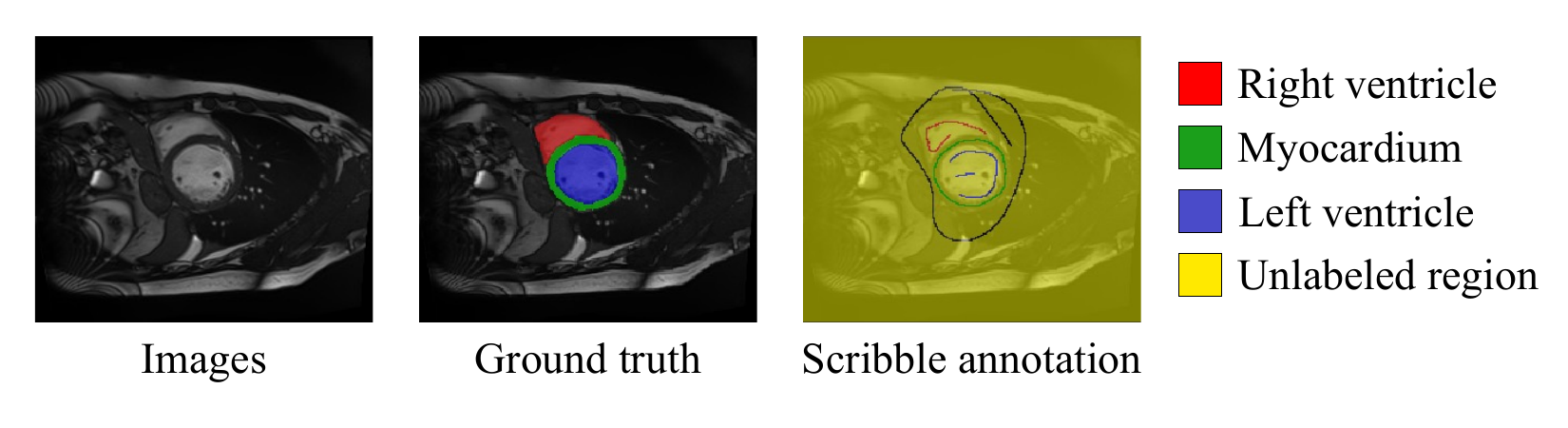}
   \end{center}
   \caption{An example slice of the ACDC dataset. }
\end{figure}

\begin{figure}[!htb]
   \begin{center}
   \includegraphics[width=0.85\textwidth]{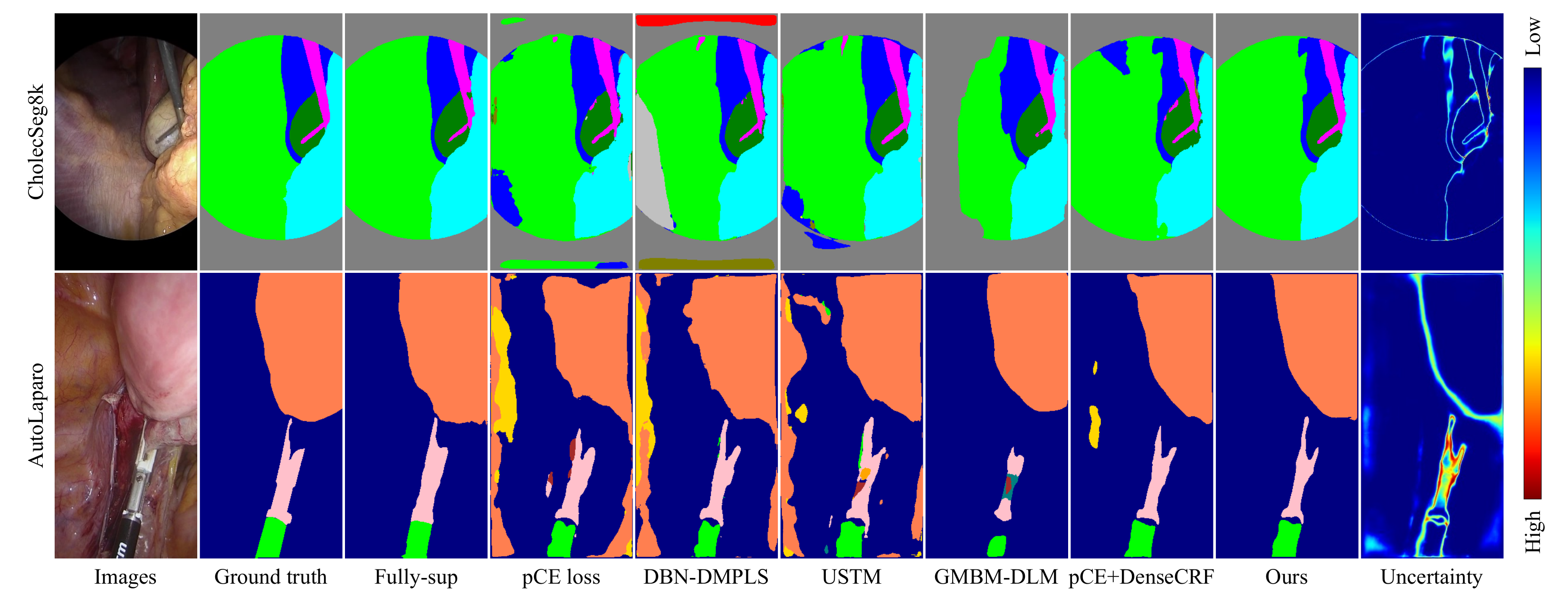}
   \end{center}
   \caption{Visualization results of various methods.}\label{visualization_results}
\end{figure}

\end{document}